\documentclass[11pt]{article}
\usepackage[T1]{fontenc}
\usepackage[utf8]{inputenc}
\usepackage[margin=1in]{geometry}
\usepackage{amsmath,amssymb,booktabs,tabularx,array}
\usepackage{graphicx,xcolor,listings,hyperref}
\hypersetup{colorlinks=true,linkcolor=blue,urlcolor=blue,citecolor=blue}
\newcommand{\step}[1]{\par\medskip\noindent\hspace*{1em}\textbf{#1.}\ }
\title{Wikidata Search Traces: A Dataset for Training Knowledge Graph Search Agents}
\author{
Mohamed Chenene\textsuperscript{1} \quad Carlos Rosas-Hinostroza\textsuperscript{1,2}\\
Anastasia Stasenko\textsuperscript{1,3} \quad Shani Evenstein Sigalov\textsuperscript{4} \quad Pierre-Carl Langlais\textsuperscript{1,5,6}\\[0.6em]
\small\textsuperscript{1}PleIAs\\
\small\textsuperscript{2}Lattice, ENS-PSL, CNRS, Sorbonne Nouvelle\\
\small\textsuperscript{3}Université Paris Dauphine-PSL\\
\small\textsuperscript{4}AI-BRIDGES, Digital Humanities Research Hub, School of Advanced Study, University of London\\
\small\textsuperscript{5}Sorbonne Center for Artificial Intelligence\\
\small\textsuperscript{6}Sciences Po M\'edialab\\
}
\date{October 2026}
\begin{document}
\maketitle
{\let\thefootnote\relax\footnotetext{Traces dataset: \url{https://huggingface.co/datasets/AI-BRIDGES/wikidata-search-traces}. RLM harness: \url{https://github.com/aibridges/RLM_Wikidata}.}}
\begin{abstract}
Wikidata is one of the largest open knowledge bases, yet answering a complex question over it still requires a SPARQL query that names the right entities and properties and chains their relations. Language models offer a natural-language alternative but answer largely from memory, which is least reliable for less prominent entities. We study agents that instead answer by exploring the graph, and argue that two obstacles limit them: the lack of training data recording how a solver explores, and interfaces that add large graph results directly to the model's context. We test three hypotheses: that the difficulty of graph search can be controlled through the structure of a question rather than only through obscure entities or wording; that much of the failure on long-horizon search comes from how retrieved evidence is managed rather than from the model itself; and that, in a suitable environment, open-weight models can match commercial closed ones. We construct multi-hop questions on a frozen Wikidata snapshot by replacing named entities with nested conditions, checking after each expansion that the target remains unique and that every new condition is necessary. We release 10,235 solving traces over single-entity and multi-hop questions, together with the recursive language model (RLM) harness that produced them, in which models batch graph calls, keep results in persistent Python state and interpret selected evidence through sub-calls. On 100 questions, the harness improves both models we ran under both interfaces compared with direct tool calling over the same functions: gpt-6-luna rises from 49 to 61 correct answers, doubling its multi-hop accuracy, and Qwen3.8-27B, an open-weight model served on a single GPU, from 60 to 74. 
\end{abstract}

\section{Introduction}\label{sec:intro}
Wikidata has become the leading infrastructure for the semantic web and for general-purpose structured open data \cite{wikidata}. It describes more than 100 million entities and is maintained by a volunteer community. Originally built for Wikimedia projects, it is now used by cultural heritage institutions, researchers, commercial services. It is also a major common ground for semantic databases in multiple industrial sectors, either through straight data integration, inspiration for data harmonization and ontology or query federation.

Conversational language models are becoming a common way to ask factual questions and rely more and more on internet or local databases. Yet, Wikidata and the wider semantic web remain poorly integrated in this emerging search environment. A language model answers mostly from what it memorized during training, which is least reliable for less prominent entities \cite{popqa,kandpal}, and it gives no account of where an answer comes from. A knowledge graph holds explicit, sourced statements, but only users who can write structured queries can combine them. A model that answers by consulting the graph rather than recalling it would bring together the accessibility of the first and the verifiability of the second.

Consulting the graph is itself difficult. To write a SPARQL query, a user must identify the relevant entities and properties, follow the direction of their relations and express joins between the conditions of the question. Natural-language interfaces have mostly framed this as semantic parsing: LC-QuAD 2.0 and KQA Pro pair questions with executable representations, and the Wikidata Query Logs dataset (WDQL) derives question--query pairs from real query logs \cite{lcquad,kqapro,wdql}. This framing assumes that a question can be mapped to its query in one step. It becomes fragile when a question describes its answer through several unnamed intermediates, for example a race held at a venue described only by its type, during a season identified through another event. The solver must then discover which entities and relations are involved, compare candidates and combine evidence gathered across many calls. The task becomes one of exploration.

Agents that interleave reasoning with retrieval are a natural fit for this task \cite{react}, but two obstacles limit them on knowledge graphs. The first concerns data. Training an agent to explore requires traces of exploration over tasks whose difficulty can be varied and whose answers can be checked. Recent synthetic datasets for search agents raise difficulty by obscuring entities, sampling rare seeds or dispersing evidence \cite{deepdive,websailor,redsearcher}. When the evidence lives on the open web, however, it is hard to certify that a question has exactly one answer and that each of its clues is needed. A frozen knowledge graph allows both checks. The second obstacle concerns the interface. One navigation step on a hub entity can return thousands of neighbours and statements. Tool-calling agents add these results to their conversation, and long-context studies show that models miss evidence depending on its position and lose accuracy as input grows \cite{lostmiddle,ruler}. 

A third question concerns openness. The strongest agents are commercial models that hide their reasoning and whose terms restrict the use of their outputs for training. For a public resource such as Wikidata, and for the institutions that rely on it, what matters is whether open models running on modest hardware can do this work.

Recursive Language Models (RLMs) offer an alternative: the model inspects an external environment through code, keeps intermediate results in persistent state and delegates the reading of selected evidence to sub-calls \cite{rlm}. Prime Intellect's nano-rlm and Prime Agent apply this approach to coding harnesses \cite{nanorlm,primeagent}. While there is a natural convergence between the concept of depth in knowledge graph exploration (through graph traversal) and recursive language model (through persistent state and sub-calls), this remains untested so far.

We formulate three hypotheses.
\begin{description}
\item[H1 (Structure).] The difficulty of a search task over a knowledge graph can be controlled through the structure of the question, meaning the number of nested dependencies to resolve and the ambiguity of each condition, provided every condition is needed to identify the answer. This would make structure a lever of difficulty alongside entity obscurity and obfuscated wording.
\item[H2 (Evidence management).] On long-horizon graph search, many failures come from how a model manages retrieved evidence rather than from missing knowledge or reasoning ability. Keeping graph results in persistent program state, which the model inspects selectively, should improve accuracy with the model held fixed, mostly on multi-hop questions.
\item[H3 (Open models).] In such an environment, an open-weight model served on a single GPU can match commercial models.
\end{description}

To test these hypotheses, we make three contributions.
\begin{enumerate}
\item \textbf{A construction method and dataset.} We grow a question graph around a known answer. An LLM selects a named entity and replaces it with a bundle of conditions drawn from the graph. Code enumerates the bundles and certifies that the graph still identifies the target alone and that removing any new condition admits another target. From seeds that vary entity type, language and prominence, we retain 10,235 solving traces over single-entity questions and multi-hop questions of several answer types.
\item \textbf{An environment.} We release the RLM harness that produced the traces, with 13 graph functions over a frozen February 2026 Wikidata snapshot. Others can replicate and extend the traces and evaluate further models in the same environment.
\item \textbf{A first evaluation.} On 100 questions, we run two models under both the harness and direct tool calling over the same functions. The harness raises gpt-6-luna from 49 to 61 correct answers and Qwen3.8-27B, served on one H100 GPU, from 60 to 74. On multi-hop questions, it takes gpt-6-luna from 9 to 19 out of 50 and Qwen3.8-27B from 20 to 28.
\end{enumerate}

\section{Related work}\label{sec:related}
\step{Questions over knowledge graphs}
LC-QuAD 2.0 generates questions paired with SPARQL queries over Wikidata and DBpedia \cite{lcquad}. KQA Pro adds explicit compositional programs covering relations, attributes, qualifiers and comparisons \cite{kqapro}, while GrailQA evaluates generalization to new combinations and unseen schema elements \cite{grailqa}. WDQL builds 335,000 question--query pairs from Wikidata Query Service logs: an agent cleans and validates logged queries and writes their corresponding questions \cite{wdql}. These datasets provide executable supervision from templates, programs or queries written by users. KGQAGen combines LLM-based question generation over Wikidata with symbolic answer verification \cite{kgqagen}. We use executable queries during construction and scoring, while the evaluated solver explores the graph through functions without receiving the answer query.

\step{Synthetic questions for search agents}
DeepDive constructs search questions from paths through a knowledge graph, with an LLM selecting successive nodes and obscuring identifying information \cite{deepdive}. WebSailor combines seed sampling, graph exploration and information obfuscation to increase uncertainty during search \cite{websailor}. REDSearcher uses graph topology and evidence dispersion to guide task synthesis \cite{redsearcher}. SearchArt describes a pipeline from seed sampling and evidence-graph construction to question writing and trajectory verification \cite{searchart}. We draw on its attention to seed exposure, graph structure and domain coverage. These approaches motivate several dimensions of task complexity; our construction focuses on nested dependencies and the contribution of individual conditions to identifying a target.

\step{Question structure and necessary conditions}
InfoSeek formulates deep research as hierarchical constraint satisfaction, combining parallel constraints with dependent sub-questions \cite{infoseek}. WebShaper represents information seeking through knowledge projections and set operations, then increases complexity by replacing a known entity with a sub-question \cite{webshaper}. We adopt this replacement as an executable graph expansion. MuSiQue composes single-hop questions into connected reasoning chains and filters examples that permit shortcuts \cite{musique}. LoHoSearch requires clues to identify one answer together and uses leave-one-out checks to test their necessity \cite{lohosearch}. We apply these checks after each expansion: the complete graph must identify the seed alone, and removing any new condition must admit another target. These guarantees concern the frozen graph; the later natural-language question still requires a fidelity check.

\step{Agents and programmatic exploration}
ReAct interleaves reasoning with actions that retrieve external information \cite{react}. Think-on-Graph guides exploration through entities and relations selected by an LLM \cite{tog}; KG-Agent combines graph tools, an executor and knowledge memory for multi-step reasoning \cite{kgagent}. CodeAct uses executable Python actions to compose operations and retain intermediate results \cite{codeact}. Our comparison uses the same graph functions through direct tool calling and through Python, so the main interface difference concerns how the model orchestrates calls and manages retrieved evidence. The budgets permit different numbers of graph calls, which we report alongside accuracy.

\step{Context management and RLM harnesses}
Lost in the Middle documents sensitivity to the position of evidence in long contexts \cite{lostmiddle}. RULER tests retrieval, multi-hop tracing and aggregation, showing that a model's advertised context window can exceed the length at which it handles these tasks reliably \cite{ruler}. RLMs move context inspection into an external environment and let the model process selected portions through recursive calls \cite{rlm}. Prime Intellect's nano-rlm implements a minimal harness with a persistent IPython environment and recursive subagents \cite{nanorlm}. Prime Agent extends this direction to a coding harness with persistent state and mechanisms for improvement across runs \cite{primeagent}. We apply RLM-style execution to knowledge-graph navigation. Our evidence-interpretation helpers receive selected data without graph access, while the main model stores and filters graph results in Python. The evaluation tests this combined interface rather than isolating the effect of recursive calls.

\section{Methodology}\label{sec:method}
\subsection{Problem definition}\label{sec:problem}
We represent a question over a knowledge graph as a set of constraints on an unknown answer. Answering the question means finding the entity that satisfies all constraints together.

\step{Representation of a question}
Let $G\subseteq E\times P\times E$ be a knowledge graph, where $E$ is the set of entities and $P$ the set of properties. A \emph{fact} $(u,p,v)\in G$ links entity $u$ to entity $v$ through property $p$; $u$ and $v$ are \emph{neighbours}.

A \emph{condition} restricts the answer through one property and a set of entities, for example ``took place at Circuit de Monaco''. Following WebShaper \cite{webshaper}, we write a condition as a \emph{knowledge projection}. For a set of entities $S\subseteq E$,
\begin{equation}
R_p(S)=\{x\in E:\exists y\in S,\ (x,p,y)\in G\}
\end{equation}
is the condition's \emph{candidate set}: the entities linked to a member of $S$ through $p$. A condition in the opposite direction uses the reversed relation, $R^{-1}_p(S)=\{x\in E:\exists y\in S,\ (y,p,x)\in G\}$. An entity satisfies several conditions when it belongs to the intersection of their candidate sets.

The \emph{candidate count} of a condition on a single entity $y$ is $|R_p(\{y\})|$: the number of entities in the whole graph that satisfy this condition alone, using the reversed projection for incoming conditions. Participation in the 24 Hours of Le Mans has a candidate count of 2,331. We use this count as a proxy for ambiguity: a condition with more candidates restricts the answer less on its own and can require the solver to combine more clues.

The \emph{target set} $T$ contains the entities satisfying all conditions on the answer. For example, ``Which race took place at Circuit de Monaco and was part of the 1986 Formula One World Championship?'' gives
\begin{equation}
T=R_{\mathrm{location}}(\{\text{Circuit de Monaco}\})
\cap R_{\mathrm{part\ of}}(\{\text{1986 season}\}),
\end{equation}
where ``1986 season'' abbreviates the championship's name. The first projection returns races at the venue; the second returns events of the season. We build each question around a known answer entity, the \emph{seed} $s$, and require $T=\{s\}$.

\step{Constructing long-horizon tasks}
To add a dependency, we turn a \emph{named} entity into a \emph{hidden} one that the solver must identify through further conditions. Two operators do this. \textbf{SELECT} chooses which named entity of the question graph to hide next. \textbf{EXPAND} replaces its singleton set with a set defined by a \emph{bundle}: one to three conditions describing that entity. The first, or \emph{root}, expansion describes the seed itself and defines $T$. Later expansions introduce the sets $V_1,V_2,\ldots$ of hidden intermediate entities.

In the Monaco example, SELECT chooses Circuit de Monaco. EXPAND describes the venue through its type and sport:
\begin{align}
V_1&=R_{\mathrm{instance\ of}}(\{\text{street circuit}\})
\cap R_{\mathrm{sport}}(\{\text{auto racing}\}),\\
T&=R_{\mathrm{location}}(V_1)
\cap R_{\mathrm{part\ of}}(\{\text{1986 season}\}).
\end{align}
The two-condition bundle defines $V_1$, the candidate venues. The question now describes the race through a venue it does not name. Repeating SELECT and EXPAND adds nested dependencies to create a long-horizon task.

\step{Certification and necessity}
Each expansion must keep the seed as the only answer, and each added condition must matter. For a proposed bundle $B$, let $T_B$ be the target set after the expansion, and $T_{B\setminus\{c\}}$ the target set after removing condition $c$ from the bundle while keeping the rest of the question graph. We require
\begin{align}
T_B&=\{s\}, &&\text{certification},\\
|T_{B\setminus\{c\}}|&\geq2\quad\text{for each }c\in B,
&&\text{necessity}.
\end{align}
\emph{Certification} checks that the complete question graph identifies the seed alone. \emph{Necessity} checks that removing any new condition admits another answer. Following LoHoSearch \cite{lohosearch}, we check necessity when adding a bundle. We also reject clues that reveal hidden names, including given-name and family-name clues, and require each condition to match at least two candidates so that no single condition identifies the entity alone.

\step{Graph representation}
We give the question graph to the LLM as a Mermaid-like text. Named entities are nodes with readable labels, hidden entities are variable nodes, and conditions are directed edges labelled with their property. The variable \texttt{?target} stands for the answer, and \texttt{?v1}, \texttt{?v2}, and so on for the hidden intermediate entities $V_1,V_2,\ldots$. After the expansion above, the Monaco question graph reads:
\begin{lstlisting}
?target --location--> ?v1
?target --part of--> 1986 Formula One World Championship
?v1 --instance of--> street circuit
?v1 --sport--> auto racing
\end{lstlisting}
Repeated variables denote the same entity. The LLM reads the dependencies from this text; code keeps the underlying entity and property identifiers to query the graph.

We use a frozen Wikidata snapshot from February 2026, with about 48 million entities (Wikidata items) and 262 million facts between them.

\subsection{Seed selection}\label{sec:seeds}
We select seeds to vary entity types, languages and prominence. We include less prominent entities to reduce reliance on answers a model may have memorized, while keeping entities with enough facts to build a question graph. We exclude Wikimedia maintenance entities (categories, templates and disambiguation pages) and entities with more than 500 incoming or 100 outgoing links. Each seed must have at least one sitelink, providing a linked Wikimedia page, and belong to a common class, such as \emph{human}. We then require an initial bundle that passes certification and necessity.

We reserve a share of seeds described only in Arabic, Russian or Chinese to vary language coverage. To balance prominence, we use the long-tail score
\begin{equation}
L(e)=\frac{1-r(d_{\mathrm{in}}(e))+r(q(e))}{2},
\end{equation}
where $r$ is the percentile midrank among eligible entities, $d_{\mathrm{in}}(e)$ the incoming degree and $q(e)$ the numeric part of the QID. Low incoming degree and a higher QID number increase the score. We sample across popular ($L(e)<.33$), moderate ($L(e)\in[.33,.66)$) and rare ($L(e)\geq.66$) seeds. We selected 2,973 seeds and attempted generation from 2,160 of them. Appendix~\ref{app:metrics} reports the studies used to assess whether a seed supports continued growth.

\subsection{Growing the question graph}
\begin{figure}[t]
    \centering
    \includegraphics[width=\linewidth]{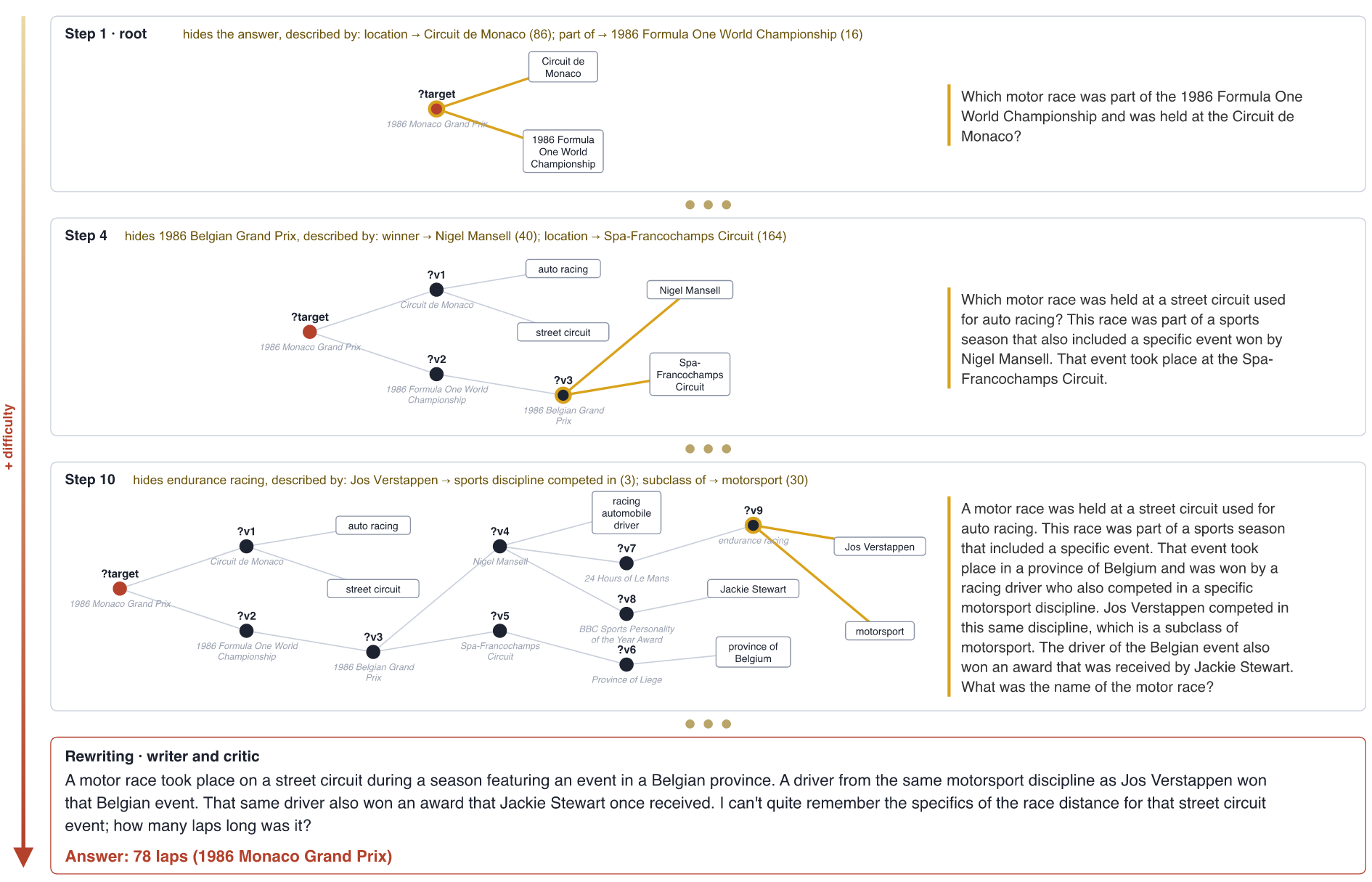}
\caption{Graph growth through SELECT and EXPAND. Successive panels show how the same target remains identifiable as named entities become variables and the graph expands. The final Monaco graph contains 16 relations after ten expansions.}
\label{fig:growth}
\end{figure}

We use an LLM to choose entities and bundles that form a coherent question from the verified candidates. Both prompts include the current graph and a \emph{working question}, a natural-language rendering of its conditions that anchors the model's choices. Figure~\ref{fig:growth} shows successive expansions of the Monaco example.

\step{SELECT}
We retain named entities with a label, a description, at most 20,000 incoming and 2,000 outgoing links, and at least two usable conditions: incoming or outgoing relations around the entity that pass filtering. We exclude entities used only as classes, the objects of instance-of or subclass-of conditions. We also check \emph{entity necessity}: replacing the name with an unconstrained variable must admit another target answer. This applies the necessity principle of Section~\ref{sec:problem} to the selected entity; later expansions can make an earlier name redundant.

For each remaining entity, we give the LLM a card containing its label, types, up to three descriptions and a relation profile. The profile groups usable conditions by direction and neighbour type, with counts per property, so the LLM can judge how to describe the entity. The LLM receives at most 20 cards and returns a candidate identifier with a reason, or rejects all candidates.

\step{EXPAND}We enumerate bundles of one to three conditions and use SPARQL to retain those passing certification and necessity (Section~\ref{sec:problem}). We rank them by decreasing size to add more joint constraints, then by decreasing sum of candidate counts within each size. This favours broader individual clues that require the solver to combine conditions.

The LLM receives up to 50 ranked bundles, each listing its conditions and candidate counts, together with the question graph, working question and selected entity's label and description. The selected entity appears as \texttt{?x}. The LLM chooses a bundle and rewrites the working question to express the updated graph. It returns a bundle identifier, a reason and the new working question without adding or editing conditions.

\step{Stopping criterion}
We repeat SELECT and EXPAND, counting the root as the first expansion. We require at least three expansions and depth three, with limits of ten expansions and depth seven. Depth is the longest shortest-path distance from the target to a named entity, reading edges in both directions. We chose these structural thresholds empirically; Appendix~\ref{app:implementation} gives implementation details.

\subsection{From graph to question}\label{sec:toquestion}
After growth, we vary what to ask about the identified target (Table~\ref{tab:questiontypes}) and read the expected answer from the graph or entity record. Sampling targets are 60\% identity, 20\% year, 10\% quantity and 10\% shared across group, count, argmax and argmin. Filtering changes the observed proportions.

\begin{table}[ht]
\centering\small
\begin{tabularx}{\linewidth}{@{}lXX@{}}
\toprule
Question type & Requested answer & Illustrative question \\
\midrule
Identity & The target entity & Which race matches this description? \\
Year & A year from a target property or relation qualifier & In which year did this race take place? \\
Quantity & A measured value and its unit & How many laps long was this race? \\
Group & All entities pointing to the target through a specified property & Which drivers participated in this race? \\
Count & The number of entities in that group & How many drivers participated in this race? \\
Argmax & The group member with the highest value or latest date & Who was the youngest participant in this race? \\
Argmin & The group member with the lowest value or earliest date & Who was the oldest participant in this race? \\
\bottomrule
\end{tabularx}
\caption{Multi-hop question types. Illustrative question endings after a description identifying a race. Group tasks use participation links; extrema compare participants' dates of birth.}
\label{tab:questiontypes}
\end{table}

A writer LLM turns the question graph into a question, in the voice of a persona drawn from Nemotron-Personas-USA and in a register set by the persona's education. Mechanical checks reject questions that contain an identifier, the name of a hidden entity or the answer. A critic LLM checks that the question states every condition, asks for the right answer and reads naturally. Appendix~\ref{app:writingprompts} gives the writer and critic instructions.

From 2,003 certified question graphs we accepted 1,498 questions, including 227 with a wording-only waiver that preserves the fidelity and answer checks. A later pass rewrote question openings, with a second LLM checking each rewrite; it changed 1,443 of the 1,498 questions and removed none.

We also generate \emph{single-entity questions}, which name the target and ask about a selected fact without growing a question graph. They use the same writer and mechanical checks, without a critic. The batch contains 11,130 accepted questions over 4,427 entities.

\subsection{Solving traces and retention}\label{sec:traces}
A \emph{solving trace} records a model answering a question in the RLM harness (Section~\ref{sec:harness}). During collection, we provide graph-derived clues to help the model navigate towards the answer and obtain successful traces for future training. Gemini-3.1-flash-lite runs at temperature zero, without thinking and within 30 turns. We retain traces whose final answer matches the expected answer and remove the collection clues from the exported prompt. Table~\ref{tab:retention} gives the counts; Appendix~\ref{app:traceprompt} summarizes the instructions.

\begin{table}[t]
\centering\small
\begin{tabular}{@{}lrr@{}}
\toprule
Stage & Multi-hop & Single-entity \\
\midrule
Attempted seeds / selected facts & 2,160 & 11,199 \\
Certified graphs & 2,003 & --- \\
Accepted questions & 1,498 & 11,130 \\
Attempted solving traces & 1,224 & 11,130 \\
Retained traces & 672 & 9,563 \\
\bottomrule
\end{tabular}
\caption{Generation and retention counts. Single-entity generation starts from selected facts. Each example counts once despite revisions or retries.}
\label{tab:retention}
\end{table}

Trace acceptance removes 552 of 1,224 attempted multi-hop questions and 1,567 of 11,130 single-entity questions. The released corpus contains 10,235 traces, with a median length of ten turns for multi-hop and one for single-entity questions. Cleaning removes 10,042 error turns and keeps 3,624; these count turns, not lost questions.

These reductions measure lost questions. They do not measure what information is lost between the question graph and its written question: the critic checks fidelity, but the batch has no human estimate of that loss. Trace filtering also shifts the distribution of question types, because the solving model completes some types more often than others.

\section{Evaluation}\label{sec:eval}
The evaluation checks that the harness used to produce the traces is a useful environment for solving the questions. It asks whether the harness helps a model compared with calling the same graph functions directly, and how far it takes an open-weight model served on a single GPU. These two questions correspond to H2 and H3 (Section~\ref{sec:intro}); the comparison between single-entity and multi-hop questions also bears indirectly on H1.

\subsection{Harness and protocol}\label{sec:harness}
We compare two interfaces to the same 13 graph functions (Table~\ref{tab:graphfunctions}). The \emph{RLM harness} gives the model a persistent Python REPL, where it can combine function calls, filter results and store intermediate data in variables. The \emph{tool-calling agent} receives the functions as tools and their outputs as conversation messages. We implement this baseline with the OpenAI Agents SDK, using its native OpenAI integration and LiteLLM for the other providers.

\begin{table}[ht]
\centering\small
\begin{tabularx}{\linewidth}{@{}p{.29\linewidth}X@{}}
\toprule
Function & Returned information \\
\midrule
\texttt{search\_entity} & Entities matching a name through labels and aliases, with the total match count \\
\texttt{search\_property} & Properties matching a name, with the total match count \\
\texttt{claims} & Statements, typed values, ranks and qualifiers; preferred statements by default when present, otherwise non-deprecated statements \\
\texttt{references} & Recorded sources for statements, optionally restricted to one property \\
\texttt{describe} & Entity types, labels, descriptions, aliases and sitelinks across six languages \\
\texttt{labels}, \texttt{label} & Available labels across the six languages, or a label in one specified language \\
\texttt{descriptions} & Available descriptions across the six languages \\
\texttt{edges} & Incoming and outgoing entity links, with property and neighbour labels; supports property filtering and pagination \\
\texttt{count\_edges} & Counts of incoming and outgoing links, optionally restricted to a property \\
\texttt{degree} & Total outgoing and incoming link counts \\
\texttt{name} & The English label of an entity or property \\
\texttt{has} & Whether an entity or property identifier exists in the graph \\
\bottomrule
\end{tabularx}
\caption{Graph functions shared by both evaluation interfaces. The six languages are English, French, German, Chinese, Arabic and Russian.}
\label{tab:graphfunctions}
\end{table}

The RLM also provides \texttt{llm\_query(text, instruction)} to interpret supplied evidence through a model sub-call, and \texttt{llm\_map(items, instruction)} to batch such requests and return an answer per item. These helpers have no graph access. The RLM submits its answer through \texttt{FINAL(value)} or \texttt{FINAL\_VAR(value)}; the agent returns structured output in the requested answer format. Both interfaces refuse graph calls using identifiers absent from the question and earlier results.

We use low reasoning effort for all models in the main evaluation. Agents allow up to 100 model calls and a \$0.50 cost cap per question, which does not apply to the locally served Qwen3.8-27B; RLM runs allow up to 100 turns and 80 sub-calls. Both interfaces truncate visible outputs at 10,000 characters. These limits do not equalize graph calls: a Python cell can execute many graph calls. The shared instruction tells models that the question has an answer and to continue until they find it.

The main evaluation contains 50 single-entity and 50 multi-hop questions built with the pipeline of Section~\ref{sec:method}. Evaluated models receive no collection clues.

We ran gpt-6-luna and Qwen3.8-27B under both interfaces, which measures the effect of the harness for each of them. GLM-5.3-flash and Gemini-3.1-flash-lite ran only as tool-calling agents. Qwen3.8-27B runs in FP8 with vLLM on one H100 GPU, with thinking enabled at low effort, temperature 1 and up to 8,192 output tokens per reply, under both interfaces. As an agent it uses vLLM's tool-call parser and returns its answer through an answer tool, because the server cannot combine tool calls with a constrained output format. gpt-6-luna and the other agents run through their providers' APIs.

Using an open-weight model also matters for the corpus. Commercial frontier models such as gpt-6-luna reason internally and return only their final reply, whereas Qwen3.8-27B returns its full reasoning, which the harness records beside the code, outputs and graph calls of each turn. Traces generated with an open model can therefore include the reasoning behind each action, for training and auditing, and can be produced locally without per-token costs or provider terms restricting their use for training.

\subsection{Results}
\begin{table}[t]
\centering\small
\textbf{Accuracy and cost: all 100 questions}\par\smallskip
\begin{tabularx}{\linewidth}{@{}Xlrrrr@{}}
\toprule
Model & Interface & Single-entity & Multi-hop & Total & Cost/question \\
\midrule
Qwen3.8-27B & RLM harness & 46/50 & 28/50 & 74/100 & \$0.04$^{*}$ \\
gpt-6-luna & RLM harness & 42/50 & 19/50 & 61/100 & \$0.013 \\
Qwen3.8-27B & tool-calling agent & 40/50 & 20/50 & 60/100 & \$0.025$^{*}$ \\
GLM-5.3-flash & tool-calling agent & 37/50 & 13/50 & 50/100 & \$0.067 \\
gpt-6-luna & tool-calling agent & 40/50 & 9/50 & 49/100 & \$0.005 \\
Gemini-3.1-flash-lite & tool-calling agent & 31/50 & 10/50 & 41/100 & \$0.19 \\
\bottomrule
\end{tabularx}
\medskip
\textbf{Execution patterns: 50 multi-hop questions}\par\smallskip
\begin{tabularx}{\linewidth}{@{}Xlrrrrr@{}}
\toprule
Model & Interface & Target reached & Graph calls & Repeats & No answer & Tokens \\
\midrule
Qwen3.8-27B & RLM harness & 39/50 & 620 & 25\% & 1/50 & 298,059 \\
gpt-6-luna & RLM harness & 35/50 & 15,575 & 37\% & 0/50 & 75,255 \\
Qwen3.8-27B & tool-calling agent & 27/50 & 45 & 7\% & 8/50 & 181,047 \\
GLM-5.3-flash & tool-calling agent & 23/50 & 50 & 6\% & 16/50 & 140,451 \\
gpt-6-luna & tool-calling agent & 17/50 & 13 & 1\% & 0/50 & 65,726 \\
Gemini-3.1-flash-lite & tool-calling agent & 30/50 & 52 & 27\% & 20/50 & 965,267 \\
\bottomrule
\end{tabularx}
\caption{Main evaluation results and execution patterns, ordered by total accuracy. gpt-6-luna and Qwen3.8-27B ran under both interfaces. Target reached counts runs whose graph calls touched the target entity, whether or not the final answer was correct. Graph calls are rounded means per run; tokens are median input plus output tokens per run. Repeats use a function and arguments already called in the same run; no answer counts runs that ended without a final answer: at the turn or cost limit or, for the eight Qwen3.8-27B agent runs, with an empty reply. $^{*}$Qwen costs estimate batched serving on one H100 through Google Colab; API costs use token prices.}
\label{tab:results}
\end{table}

\step{The harness improves results}
The two best results both come from the RLM harness (Table~\ref{tab:results}), and it improves both models we ran under both interfaces. gpt-6-luna rises from 49 to 61 correct answers and Qwen3.8-27B from 60 to 74. For gpt-6-luna the gain is almost entirely on multi-hop questions, 19 against 9; for Qwen3.8-27B it is 28 against 20 on multi-hop questions and 46 against 40 on single-entity ones. On multi-hop questions, the harness also takes each model to the target entity more often: 35 runs against 17 for gpt-6-luna, 39 against 27 for Qwen3.8-27B. Once there, each model answers correctly at about the same rate under both interfaces, about one half for gpt-6-luna and three quarters for Qwen3.8-27B. The harness therefore helps by carrying the search further through the chain of conditions. It also brings an open model to a useful level: served on a single GPU at an estimated \$0.04 per question, Qwen3.8-27B answers most of the questions.

\step{Where the gain comes from}
Code lets the model explore broadly without placing each result in its context. This is the core principle of an RLM: the model treats the graph as an external environment, explores it from code and keeps intermediate results in persistent state, so its context holds only what it chooses to print. Through the harness, gpt-6-luna makes 15,575 graph calls per multi-hop run against 13 as an agent, at similar token use (75,255 against 65,726), and Qwen3.8-27B makes 620 against 45. Zhang et al.\ observe that this REPL environment alone captures much of an RLM's benefit, with recursive sub-calls adding value on information-dense tasks \cite{rlm}. Qwen3.8-27B persists longer, with a median of 24 turns and 298,059 tokens per multi-hop question against 13 turns and 75,255 tokens for gpt-6-luna. Among the agents, GLM-5.3-flash and Gemini-3.1-flash-lite exhaust their turn or cost limits in 16 and 20 multi-hop runs, Qwen3.8-27B ends 8 runs with an empty reply, and gpt-6-luna always answers, after a median of 12 calls.

\section{Limitations and future work}\label{sec:limits}
\step{Training smaller models}
The traces are meant for training. The next step is to fine-tune a smaller open model, around 12B parameters, on them, with one sample per harness turn as in Zhang et al., where training Qwen3-8B on 1,000 filtered trajectories improved it by a median 28.3\% \cite{rlm}. Evaluating it in the same harness, against its base model, would test whether the traces make a model smaller than 27B usable on these questions, extending H3 to smaller models. The traces follow graph-derived clues (Section~\ref{sec:traces}), so they show more direct routes than an unguided solver would take. They were collected without thinking; generating further traces with an open model in the same harness would add its full reasoning to each turn.

\step{Evaluation scope}
These first results point in a consistent direction: with the model held fixed, the harness improves accuracy for both models we ran under both interfaces, and it lets an open-weight model served on one GPU answer most of the questions. Broader testing would confirm this direction. Running GLM-5.3-flash and Gemini-3.1-flash-lite in the harness as well, and repeating runs on a larger question set, would show how far the gain extends across models. Matching graph-call budgets and enabling or disabling the sub-call helpers would show which parts of the harness contribute most.

\step{Training data}
We generated solving traces with graph-derived clues to the solution path in the model's prompt to help it reach the answer (Section~\ref{sec:traces}). The published prompts omit these clues, but collection favours direct routes with few dead ends. Each question graph also identifies one target, and each question has a fixed expected answer. The dataset therefore provides no examples that teach a model to report several valid targets or no answer.

We can create adversarial questions by relaxing or altering constraints to obtain graphs with multiple solutions or none. Once certification finds at least two targets, we would retrieve the complete answer set for exact scoring and ask the model to return all targets or flag ambiguity. For verified empty sets, we would require \texttt{NO ANSWER}. These tasks also require changing the instruction that every question has an answer.

Our representation combines constraints through intersections. Extending it to unions, exclusions and other logical structures could broaden the reasoning tasks while preserving executable queries and exact answer checks.

\step{Harness}
We should also improve the harness by reusing graph results, detecting repeated searches without progress and requiring answer verification before stopping. Batched helper calls and concise evidence summaries can control context growth. We would evaluate these changes with matched budgets, measuring accuracy, graph calls, tokens and stopping behaviour on tasks of varying difficulty.

\section{Conclusion}\label{sec:conclusion}
We started from a tension between two ways of accessing factual knowledge. Language models answer fluently from memory but least reliably on less prominent entities. A knowledge graph holds explicit, sourced statements that only structured queries can reach. We studied agents that answer by exploring Wikidata and identified two obstacles: the lack of training data recording exploration, and interfaces that flood the model's context with graph results. To address them, we released 10,235 solving traces built by growing certified question graphs around fixed answers, together with the RLM harness that produced them.

Our evidence bears unevenly on the three hypotheses.

\step{H1 (structure)}
Certification and necessity give a checkable way to grow questions whose conditions must all be used. Multi-hop questions are markedly harder than single-entity ones in every configuration we tested: 9 to 28 correct out of 50, against 31 to 46. This gap also reflects other differences between the two sets, so it does not yet show that difficulty follows the depth or number of expansions. Showing that requires an analysis by graph structure and a human estimate of how faithfully written questions preserve their graphs.

\step{H2 (evidence management)}
With the model held fixed, the harness improves both models we ran under both interfaces: gpt-6-luna from 49 to 61 and Qwen3.8-27B from 60 to 74. On multi-hop questions it doubles gpt-6-luna's accuracy (9 to 19 out of 50) and raises Qwen3.8-27B's from 20 to 28, while each model reaches the target entity more often (17 to 35 and 27 to 39). Once the target is reached, each model's share of correct answers stays about the same under both interfaces. The gain therefore comes from carrying the search further, not from better interpretation at the end. This is consistent with evidence management, rather than the model, limiting long-horizon search; it holds for two models, under unmatched graph-call budgets.

\step{H3 (open models)}
An open-weight model served on a single GPU gives the highest score of the evaluation, 74 out of 100, and scores above gpt-6-luna under both interfaces: 74 against 61 in the harness, 60 against 49 as an agent. Whether the traces pass this ability on to smaller models remains untested.

\medskip
These results suggest that, for structured public knowledge, part of the capability lies in the environment around the model: in how it is allowed to explore, store and verify. This matters for infrastructures such as Wikidata, where answers should be traceable to statements and their sources, and where institutions and researchers need tools they can run, inspect and adapt. We release the traces, the construction method and the harness for this purpose. 

The next steps follow from the open hypotheses: training smaller open models on the traces and evaluating them in the same harness, analysing accuracy by graph structure, adding questions with several valid answers or none, and comparing interfaces with matched budgets, with sub-calls enabled and disabled. 

More broadly, our future work aims to investigate Recursive Language Models as a generalized approach for semantic data search beyond Wikidata. Interaction between different sources and ontologies (or "database federation") has long been a leading premises of semantic web and advanced knowledge management, constantly undermined by data frictions and high associated maintenance costs. Already back in 2001, Berners-Lee and colleagues predicted that "the Semantic Web will bring structure to the meaningful content of Web pages, creating an environment where software agents roaming from page to page can readily carry out sophisticated tasks for users'' \cite{berners-lee}. Different forms of agents, based on generative AI, finally took hold of the web. And yet, building the proper interactive surface and internal capabilities to properly deal with the complexity of data remain still to this day an open program.

\section*{Acknowledgments}
This work was carried out as part of \href{https://ai-bridges.org}{AI-BRIDGES}, a project funded by the European Commission through a Marie Skłodowska-Curie Postdoctoral Research Fellowship (Grant ID: 101203096), hosted at the Digital Humanities Research Hub, School of Advanced Study, University of London. The work was also funded by Wikimedia Switzerland (WMCH) and conducted in close collaboration with Wikimedia Deutschland (WMDE). Harness development and evaluations were supported by compute grant n°AD011014736R2 on the Jean Zay H100 cluster.

\clearpage\appendix
\section{Seed-selection studies}\label{app:metrics}
We tested which seed measures predict successful graph growth. The first study used 50 seeds with three runs per seed; success meant that at least one run reached three expansions and depth three. We replayed the trajectories to inspect where growth stopped. The second study used 100 seeds (33 popular, 34 moderate and 33 rare), with one run per seed and the same stopping criterion.

\step{Measures and results}
A usable entity passes SELECT filtering and has a valid expansion bundle. A describable neighbour has a valid bundle; a usable condition is an incoming or outgoing relation retained after filtering. AUC measures how often a score ranks a successful seed above a failed seed, with half credit for ties: .5 is chance ranking and 1 is perfect separation. Table~\ref{tab:seedstudies} gives AUCs and the reported 95\% intervals. The two studies use different success endpoints, so their values describe separate probes.

\begin{table}[ht]
\centering\small
\begin{tabularx}{\linewidth}{@{}Xll@{}}
\toprule
Measure & 50-seed study & 100-seed study \\
\midrule
Usable entities after root & .88 [.78, .96] & .81 [.68, .92] \\
Share of root bundles opening a three-hop chain & .82 [.70, .93] & --- \\
Share of root bundles leaving a usable entity & .80 [.66, .91] & --- \\
Number of three-hop chains & .78 [.64, .90] & --- \\
Outgoing degree & --- & .72 [.61, .82] \\
Number of valid root bundles & --- & .71 [.58, .83] \\
Usable-condition count & .66 [.49, .80] & .69 [.57, .81] \\
Describable-neighbour count & .67 [.55, .78] & .53 [.44, .62] \\
Largest valid bundle size & .63 [.51, .76] & --- \\
\bottomrule
\end{tabularx}
\caption{Seed-level growth predictors. The first row uses the accepted root expansion; root-bundle and chain measures can be computed before a run. A dash indicates a measure not reported in that study.}
\label{tab:seedstudies}
\end{table}

The ability to continue after the root was more informative than the number of available clues. In the trajectory study, none of the 70 runs with no usable entity after the root succeeded. In the 100-seed study, 13 of 21 failed runs had no usable entity after the root, while no successful run had this problem. These results motivated screening seeds for a starting bundle that permits further expansion. They measure growth, not question quality; the samples are small and the first study used a more restrictive certification executor.

\section{SELECT and EXPAND}\label{app:implementation}\label{app:prompts}
\step{Candidate construction}
We retain conditions with an allowed property, a labelled neighbour, at least two candidates and support beyond deprecated-only statements. We exclude given-name and family-name clues, cap the pool at 24 conditions and spread it across properties. Exact target queries stop after finding a second answer. We enumerate singletons, pairs and triples: a certified smaller bundle needs no added condition, and a larger bundle is retained only if each immediate sub-bundle admits another answer. This checks certification and necessity during enumeration.

SELECT requires at least two usable conditions and a non-redundant named entity. If more than 20 entities qualify, we keep those with the most usable conditions. Each relation profile shows up to eight direction--neighbour-type groups and the three most frequent properties in each group. EXPAND ranks bundles by size, then by the sum of individual candidate counts, and presents up to 50 per shortlist. Enumeration allows 600 certification queries per step and 30 minutes per seed.

\step{Entity-level tests}\label{app:entitymetrics}
We tested entity measures against successful expansion in the 100-seed probe and in the replayed trajectories (446 attempted entities). Table~\ref{tab:entitystudies} separates static measures from checks of an entity's contribution to the current question graph.

\begin{table}[ht]
\centering\small
\begin{tabularx}{\linewidth}{@{}>{\raggedright\arraybackslash}p{.31\linewidth}>{\raggedright\arraybackslash}Xr@{}}
\toprule
Measure & Definition & AUC \\
\midrule
Incoming/outgoing degree & Links pointing to/from the entity & .53/.56 \\
Usable-condition count & Retained incoming and outgoing relations & .54 \\
Largest valid bundle size & Most conditions in a valid bundle & .53 \\
Describable-neighbour count & Neighbours with a valid bundle & .54 \\
\midrule
Entity-necessity query resolvability & Whether the target query resolves after replacing the name with an unconstrained variable & .78 \\
Entity-necessity target-answer count & Number of target answers after that replacement; one means the name is redundant & .97$^{*}$ \\
\bottomrule
\end{tabularx}
\caption{Predictors of successful entity expansion. Static measures use the 100-seed probe; entity-necessity measures use replayed trajectories. $^{*}$Computed only for resolvable queries.}
\label{tab:entitystudies}
\end{table}

Static measures were close to chance, whereas the entity's role in the current graph was informative. Among 647 selected entities passing entity necessity in the 100-seed probe, 632 expanded successfully (97.7\%). We therefore use entity necessity as a gate and give the LLM relation profiles instead of static degree scores. These observations motivate the design; they do not isolate its effect through a controlled ablation.

\step{SELECT prompt}\label{app:candidateexamples}
The following templates condense the two separate LLM calls. Generation uses Gemini-3-flash-preview with low reasoning effort. Placeholders indicate supplied values.
\begin{lstlisting}
CONTEXT
Build a coherent research question around one hidden answer.
Choose only supplied candidates; do not invent or edit facts.
Never reveal private names, identifiers or answer values.

INPUTS
private_target: <label and description; never expose>
question_graph: <Mermaid-like labelled relations>
working_question: <current natural-language question>
remaining_expansions: <remaining expansion budget>
candidates: <up to 20 cards>
  Candidate <id>: <label> (<types>)
    <language>: <description> | ...
    <source type> --> <neighbour type>: <property> (<local count>)

TASK
Choose an entity whose relations support a coherent next
subproblem. Avoid repetitive, arbitrary or revealing clues.
Reject all candidates if none offers a useful continuation.

OUTPUT
{"candidate_id": <id>, "reject": false, "reason": "..."}
To reject: set candidate_id to null and reject to true.
\end{lstlisting}

\step{EXPAND prompt}
The Monaco bundle illustrates the candidate format.
\begin{lstlisting}
CONTEXT
Replace the selected entity with one supplied, verified bundle.
Each bundle preserves the unique answer and condition necessity.
Do not invent or edit facts or reveal private names and answers.

INPUTS
private_target: <label and description; never expose>
question_graph: <selected entity replaced by ?x>
working_question: <current natural-language question>
selected_entity: <label and description>
bundles: <up to 50 verified, ranked bundles>
  Candidate <id>:
    ?x --instance of--> street circuit [candidate_count=149]
    ?x --sport--> auto racing [candidate_count=16593]

TASK
Choose conditions that fit together and keep the question
coherent. Prefer broad individual clues that require combining.
Rewrite the entire working question to express the updated graph,
without naming ?x or adding facts. Keep candidate counts private.

OUTPUT
{"candidate_id": <id>, "reject": false, "reason": "...",
 "working_question": "<complete rewritten question>"}
To reject: set candidate_id to null, reject to true and
working_question to an empty string; explain why in reason.
\end{lstlisting}

A local profile count describes retained relations around one node: in the Le Mans card, \texttt{participant in (4)} means four retained incoming relations from human neighbours. A candidate count describes matches in the entire graph: participation in Le Mans matches 2,331 entities. For the two-condition circuit bundle, the ranking sum is $149+16{,}593=16{,}742$; it is not the size of the intersection.

\step{Monaco progression}\label{app:monaco}
Table~\ref{tab:monacohistory} shows how ten expansions grow the running example to 16 relations. Each listed bundle passed certification and necessity.
\begin{table}[ht]
\small\centering
\begin{tabularx}{\linewidth}{@{}r p{.25\linewidth}X@{}}
\toprule
Step & Hidden entity & Added conditions \\
\midrule
1 & Seed race & Location: Circuit de Monaco; part of: 1986 championship \\
2 & Circuit de Monaco & Instance of: street circuit; sport: auto racing \\
3 & 1986 season & Has part: 1986 Belgian Grand Prix \\
4 & Belgian event & Winner: Nigel Mansell; location: Spa-Francorchamps \\
5 & Nigel Mansell & Occupation: racing driver; participant in: Le Mans; award: BBC Sports Personality of the Year \\
6 & Spa-Francorchamps & Located in: Province of Li\`ege \\
7 & Province of Li\`ege & Instance of: province of Belgium \\
8 & Le Mans & Sport: endurance racing \\
9 & BBC award & Winner: Jackie Stewart \\
10 & Endurance racing & Jos Verstappen competed in it; subclass of: motorsport \\
\bottomrule
\end{tabularx}
\caption{Accepted Monaco expansions. The first describes the seed; subsequent steps replace named entities with further conditions.}
\label{tab:monacohistory}
\end{table}

The written question asks for the race distance:
\begin{quote}\small
A motor race took place on a street circuit during a season featuring an event in a Belgian province. A driver from the same motorsport discipline as Jos Verstappen won that Belgian event. That same driver also won an award that Jackie Stewart once received. I can't quite remember the specifics of the race distance for that street circuit event; how many laps long was it?
\end{quote}
The expected answer is 78 laps. The graph identifies the race; its event-distance statement supplies the quantity.

\section{Question writing and criticism}\label{app:writingprompts}
The writer expresses the certified graph as a question; the critic checks it and can request one revision. A wording-only waiver preserves the fidelity and answer checks. The condensed templates omit the question-type definitions in Table~\ref{tab:questiontypes}. Generation uses Gemini-3-flash-preview, with medium reasoning effort for the critic.

\step{Writer prompt}
\begin{lstlisting}
INPUTS
canonical_claim: <directed relations, with shared variables>
working_question: <current draft>
question_type: <identity/year/quantity/group/count/argmax/argmin>
asked_about: <requested property, relation date or group>
private: <hidden target and expected answer>
asked_by: <persona>
style: <register and writing mode>

TASK
Express every relation, its direction and shared hidden entities.
Ask exactly for the requested answer; add no unsupported facts.
Do not expose hidden names, identifiers, variables or answers.
Use the persona's register and natural relation wording.
Keep a named starting entity and end with a direct question.

OUTPUT
{"question": "..."}
\end{lstlisting}

\step{Critic prompt}
\begin{lstlisting}
INPUTS
canonical_claim: <validated, numbered relations>
question: <writer's proposed question>
question_type: <requested answer type>
asked_about: <requested property, relation date or group>
private: <hidden target and expected answer>
asked_by: <persona>
style: <register and writing mode>

TASK
Check fidelity, answer leakage, shortcuts, the requested answer
and naturalness. Quote evidence for each issue. Propose only
rephrasing or reorganization of facts already in the graph.

OUTPUT
Accept: {"decision": "accept", "issues": [],
         "revision_instructions": []}
Revise: {"decision": "revise", "issues": [{"type": "...",
         "atom_ids": ["A1"], "evidence": "...", "reason": "..."}],
         "revision_instructions": ["..."]}
\end{lstlisting}

For single-entity questions, the writer receives the entity, selected fact, private answer, persona, style and question form. It names the entity and preserves the statement context when asking for a qualifier or source. Mechanical checks allow one revision; these questions use no critic.

\section{Solving-trace instructions}\label{app:traceprompt}
We collect traces with Gemini-3.1-flash-lite at temperature zero and no thinking, using the graph functions in Table~\ref{tab:graphfunctions}. Additional graph-derived clues support navigation during collection; the exported prompt omits them. The condensed instructions are:
\begin{lstlisting}
Answer the supplied question through the graph functions in a
persistent Python REPL. Use one Python block per turn.
Keep intermediate results in variables; print useful summaries.
Reuse earlier results. Do not invent identifiers or facts.
Use llm_query or llm_map only to interpret supplied evidence;
batch related items and validate the returned answers in code.
Finish with FINAL(value) or FINAL_VAR(value) in the requested format.

LIMITS
30 turns; up to 80 helper calls; visible output capped at 10,000
characters. Graph reads are measured but not capped.

TASK INPUTS
question: <natural-language question>
answer_format: <typed answer schema>
collection_clues: <graph-derived clues; no answer supplied>
\end{lstlisting}
We retain exact-answer traces and prune error turns when later turns do not depend on them. These are assisted collection traces for future training; evaluation uses no collection clues.

\end{document}